\documentclass[10pt, twocolumn]{article}

\pdfoutput=1

\usepackage{graphicx}
\usepackage{amsmath}
\usepackage{amsfonts}
\usepackage[titletoc, toc, title]{appendix}
\usepackage{authblk}

\begin{document}

\title{Deep Embedding Forest: Forest-based Serving \\with Deep Embedding Features}

\author[*]{J. Zhu}
\author[*]{Y. Shan}
\author[*]{JC Mao}
\author[**]{D. Yu}
\author[***]{H. Rahmanian}
\author[*]{Y. Zhang}
\affil[*]{Bing Ads of AI \& Research Group, Microsoft Corp.}
\affil[**]{MSR of AI \& Research Group, Microsoft Corp.}
\affil[***]{Department of Computer Science, UCSC}

\maketitle

\begin{abstract}
Deep Neural Networks (DNN) have demonstrated superior ability to extract high level embedding vectors from low level features.  Despite the success, the serving time is still the bottleneck due to expensive run-time computation of multiple layers of dense matrices.  GPGPU, FPGA, or ASIC-based serving systems require additional hardware that are not in the mainstream design of most commercial applications.  In contrast, tree or forest-based models are widely adopted because of low serving cost, but heavily depend on carefully engineered features.  This work proposes a Deep Embedding Forest model that benefits from the best of both worlds.  The model consists of a number of embedding layers and a forest/tree layer.  The former maps high dimensional (hundreds of thousands to millions) and heterogeneous low-level features to the lower dimensional (thousands) vectors, and the latter ensures fast serving.

Built on top of a representative DNN model called Deep Crossing ~\cite{ying2016DeepCrossing}, and two forest/tree-based models including XGBoost and LightGBM, a two-step Deep Embedding Forest algorithm is demonstrated to achieve on-par or slightly better performance as compared with the DNN counterpart, with only a fraction of  serving time on conventional hardware.  After comparing with a joint optimization algorithm called partial fuzzification, also proposed in this paper, it is concluded that the two-step Deep Embedding Forest has achieved near optimal performance. Experiments based on large scale data sets (up to $1$ billion samples) from a major sponsored search engine proves the efficacy of the proposed model.

\end{abstract}

\section{Introduction}
Well-abstracted features are known to be crucial for developing good machine learning models, but feature engineering by human usually takes a large amount of work and needs expert domain knowledge during a traditional machine learning process.  DNNs have been used as a powerful machine learning tool in both industry and research for its ability of automatic feature engineering on various kinds of raw data including but not limited to speech, text or image sources without acquiring domain expertises~\cite{lecun2015deep, seide2011feature, ying2016DeepCrossing, he2016deep}.  

With the support of hardware acceleration platforms such as clusters of general-purpose graphics processing units (GPGPUs), field programmable gate arrays (FPGAs) or ASIC-based serving systems~\cite{chetlur2014cudnn, CNTK2014, lacey2016deep}, DNNs are capable of training on billions of data with scalability.  However, DNNs are still expensive for online serving due to the fact that most of the commercial platforms are central processing units (CPUs) based with limited applications of these acceleration hardwares. 

Tree-based models such as random forests (RFs) and gradient boosted decision trees (GBDTs), on the other hand, with their run-time efficiency and good performance, are currently popular production models in large scale applications.  However, the construction of strong abstracted features to make the raw data meaningful for tree-based models requires in depth domain knowledge and is often time-consuming.     

This work proposes a hybrid model that can carry on the performance of DNNs to run-time serving with speed comparable to forest/tree-based models.

The paper is arranged in the following structure.  Sec.~\ref{sponsored_search} sets up the context of the application domain where the Deep Embedding Forest is developed. Sec.~\ref{representation} gives examples of features in the context of the sponsored search.  Sec.~\ref{sec:problemstatement} provides a formal statement of the problem by introducing the model architecture, components, and design principles.  Sec.~\ref{sec:methodology} describes the training methodology that involves the initialization of the embedding layers, the training of the forest layer, and the joint optimizationa through a partial fuzzification algorithm.  Sec.~\ref{sec:experimentresults} presents experiment results with data sets of various sizes up to 1 billion samples.  It also demonstrates the model's ability of working with different kind of forest/tree-based models.  Sec.~\ref{sec:relatedwork} elaborates on the relationships of the Deep Embedding Forest model with a number of related work.  Sec.~\ref{sec:conclusionfuturework} concludes the paper and points out the future directions.  The appendix provides the implementation details of the partial fuzzification algorithm.

\section{Sponsored Search}
\label{sponsored_search}
Deep Embedding Forest is discussed in the context of \emph{sponsored search} of a major search engine.  Readers can refer to \cite{edelman2005internet} for an overview on this subject.  In brief, sponsored search is responsible for showing ads alongside organic search results.  There are three major agents in the ecosystem: the user, the advertiser, and the search platform.  The goal of the platform is to show the user the advertisement that best matches the user's intent, which was expressed mainly through a specific query.  Below are the concepts key to the discussion that follows.

\begin{description}
  \setlength\itemsep{0em}
	\item[Query:] A text string a user types into the search box
	\item[Keyword:] A text string related to a product, specified by an advertiser to match a user query
	\item[Title:] The title of a sponsored advertisement (referred to as ``an ad'' hereafter), specified by an advertiser to capture a user's attention
	\item[Landing page:] A product's web site a user reaches when the corresponding ad is clicked by a user
	\item[Match type:] An option given to the advertiser on how closely the keyword should be matched by a user query, usually one of four kinds: \emph{exact, phrase, broad} and \emph{contextual}
	\item[Campaign:] A set of ads that share the same settings such as budget and location targeting, often used to organize products into categories
	\item[Impression:]  An instance of an ad being displayed to a user.  An impression is usually logged with other information available at run-time
	\item[Click:] An indication of whether an impression was clicked by a user.  A click is usually logged with other information available at the run-time
	\item[Click through rate:] Total number of clicks over total number of impressions
	\item[Click Prediction:] A critical model of the platform that predicts the likelihood a user clicks on a given ad for a given query
\end{description}

Sponsored search is only one kind of machine learning application.  However, given the richness of the problem space, the various types of features, and the sheer volume of data, we think our results can be generalized to other applications with similar scale.
\section{Feature Representation}
\label{representation}
This section provides example features used in the prediction models of sponsored search.  The features in Table~\ref{tab:ExampleOfIndividualFeatures} are available during run-time when an ad is displayed (an impression).  They are also available in offline logs for model training.

Each feature is represented as a vector.  Text features such as a query are converted into tri-letter grams with  $49,292$ dimensions as in~\cite{DSSM:13}\footnote{Unlike one-hot vectors, tri-letter grams are usually multi-hot and have integer values larger than $1$ on non-zero elements}.  Categorical input such as \verb|MatchType| is represented by a one-hot vector, where exact match (see Sec.~\ref{sponsored_search}) is $[1,0,0,0]$, phrase match is $[0,1,0,0]$, and so on.

There are usually millions of campaigns in a sponsored search system.  Instead of converting campaign ids into a one-hot vector with millions of dimensions, a pair of companion features is used.  Specifically, \verb|CampaignID| is a one-hot representation consisting only of the top $10,000$ campaigns with the highest number of clicks.  The $10,000^{th}$ slot (index starts from $0$) is saved for all the remaining campaigns.  Other campaigns are covered by \verb|CampaignIDCount|, which is a numerical feature that stores per campaign statistics such as \emph{click through rate}.  Such features will be referred as a \emph{counting feature} in the following discussions.  All the features introduced above are sparse features except the counting features.
\begin{table}
	\centering
		\begin{tabular}{ l | l | l }
			\hline			
			Feature name & Type & Dimension \\ \hline\hline
			Query & Text & 49,292 \\ \hline
			Keyword & Text & 49,292 \\ \hline
			Title & Text & 49,292 \\ \hline
			MatchType & Category & 4 \\ \hline
			CampaignID & ID & 10,001 \\ \hline
			CampaignIDCount & Numerical & 5 \\
			\hline  
		\end{tabular}
	\caption{Examples of heterogeneous and high dimensional features used in typical applications of sponsored search}
	\label{tab:ExampleOfIndividualFeatures}
\end{table}
\section{Problem Statement}
\label{sec:problemstatement}
\begin{figure}
	\centering
		\includegraphics[scale=0.6]{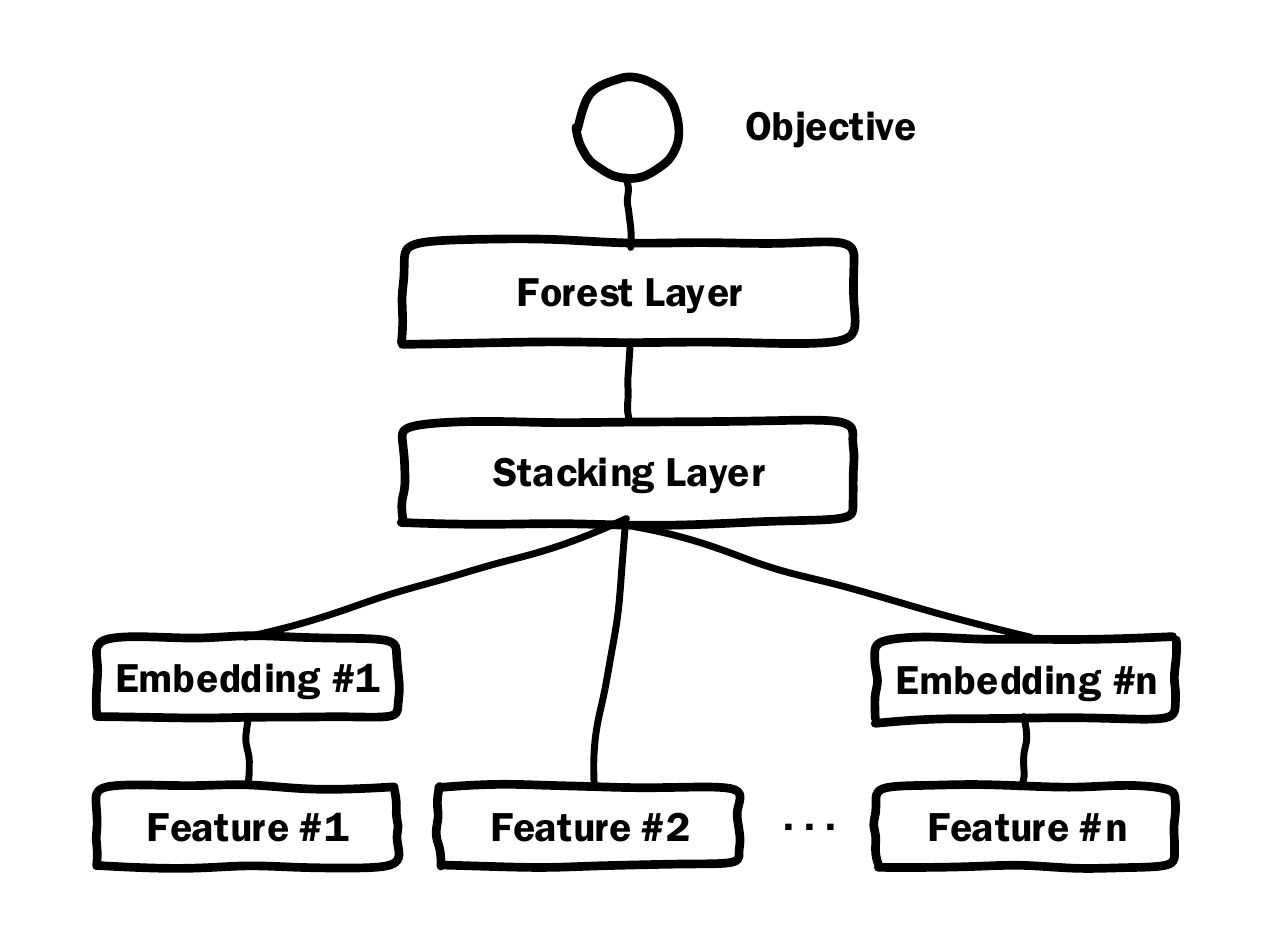}
	\caption{The Deep Embedding Forest model}
	\label{fig:defmodel}
\end{figure}

The goal is to construct a model with the structure in Fig.~\ref{fig:defmodel} that consists of feature inputs, the embedding layers, the stacking layer, the forest layer, and the objective function.  The model will be referred as \emph{DEF} or \emph{the DEF model} in the following discussions.
\subsection{Model Components}
\label{sec:modelcomponents}
DEF allows low level features of different natures including sparse one-hot or categorical features, and dense numerical features.  These features can be extracted from text, speech, or image sources.

An embedding layer maps low level features to a different feature space.  It has a single layer of neural network with the general form of:
\begin{equation}
	\tilde{\mathbf{y}}_j = g(\mathbf{W}_j\mathbf{x}_j+\mathbf{b}_j),
	\label{equ:embedding}
\end{equation}
where $j$ is the index to the individual features $\mathbf{x}_j\in \mathbb{R}^{n_j}$, $\mathbf{W}_j$ is an $m_j\times n_j$ matrix, $\mathbf{b}\in \mathbb{R}^{m_j}$ is the vector of the bias terms, $\tilde{\mathbf{y}}_j$ is the embedding vector, and $g(\cdot)$ is a non-linear activation function such as ReLU, sigmoid, or tanh.  When $m_j < n_j$, embedding is used to reduce the dimensionality of the input vector and construct high level features.

The stacking layer concatenates the embedding features into one vector as the following:
\begin{equation}
	\tilde{\mathbf{y}} = [\tilde{\mathbf{y}}_0, \tilde{\mathbf{y}}_1, \cdots, \tilde{\mathbf{y}}_{K-1}],
    \label{equ:vectorstacking}
\end{equation}
where $K$ is the number of individual features. Note that features with low dimensionality can be stacked without embedding.  An example is \verb|Feature #2| in Fig.~\ref{fig:defmodel}.

The stacking vector is sent as input to the forest layer, which is represented as $\mathcal{F}(\Psi, \Theta, \Pi)$, where $\Psi$ defines the number of trees in the forest and the corresponding structure, $\Theta$ is the parameter set of the routing functions on the decision nodes, and $\Pi$ is the parameter set of the distribution functions on leaf nodes.

DEF allows objective functions of various types including but not limited to classification, regression, and ranking.   

\subsection{DEF Properties}
\label{subsec:defproperties}
When designed and implemented properly, DEF is expected to possess two major advantages enabled by the unique model structure. 

The first is to minimize the effort of manual feature engineering.  It is usually a challenge to forest/tree-based models to handle low level features exceeding tens of thousands dimensions.  The embedding layers can comfortably operate on dimensions $10$ or $100$ times higher, and automatically generate high level embedding features to the size manageable by the forest/tree-based models.  The embedding layers are also trained together with the rest of the model using the same objective function.  As a result, they are more adapted to the applications as compared with stand-along embedding technologies such as Word2Vec~\cite{W2V:13}.

The second is to minimize the run-time latency.  Based on the structure in Fig.~\ref{fig:defmodel}, the serving time per sample is determined by the embedding time $T_1$ and prediction\footnote{The word \emph{prediction} in this context refers to a general scoring operation that is applicable to classification, regression, ranking, and so on} time $T_2$.  $T_1$, which is the run-time of a single layer of neurons, makes a small fraction of the total in a typical DNN with multiple deep layers. $T_1$ is zero for dense numerical features when they are stacked without embedding.  The complexity of embedding a sparse feature is $O(n_{\bar{0}} n_e)$, where $n_{\bar{0}}$ is the number of non-zero elements, and $n_e$ is the dimension of the corresponding embedding vector.  As an example, a sparse tri-letter gram~\cite{DSSM:13} has around $50 K$ dimensions but usually with $n_{\bar{0}}\leq 100$.  For a typical $n_e$ between $128$ and $256$, the run-time cost of an embedding layer of a sparse feature is negligible.  

The prediction time $T_2$ is a function of $n_t d_t n_f$, where $n_t$ is the number of trees, $d_t$ is the average depth of the trees in the forest, and $n_f$ is the total number of feature dimensions the decision or routing function depends on at each internal (or non-leaf) node.  DEF uses decision nodes that rely on only one feature dimension to ensure serving speed.  $T_2$ is then proportional to $n_t d_t$, which is independent of the dimensions of the stacking vector\footnote{To be more precise, the type of DEF that possesses such property is called a Two-Step DEF, as defined later in Sec.~\ref{sec:twostepandthreestep}}.  This is much cheaper than a typical DNN with multiple layers of neurons. 

 
\section{Training Methodology}
\label{sec:methodology}
Training the DEF model requires optimization of the objective function w.r.t. $\{\mathbf{W}_j\}$, $\{\mathbf{b}_j\}$, $\Psi$, $\Theta$, and $\Pi$ (see definitions in Sec.~\ref{sec:modelcomponents}).  It involves three steps detailed in the following sub-sections.

\subsection{Initialize Embedding Layers with Deep Crossing}
\label{DEFStep1}
\begin{figure}
	\centering
		\includegraphics[scale=0.6]{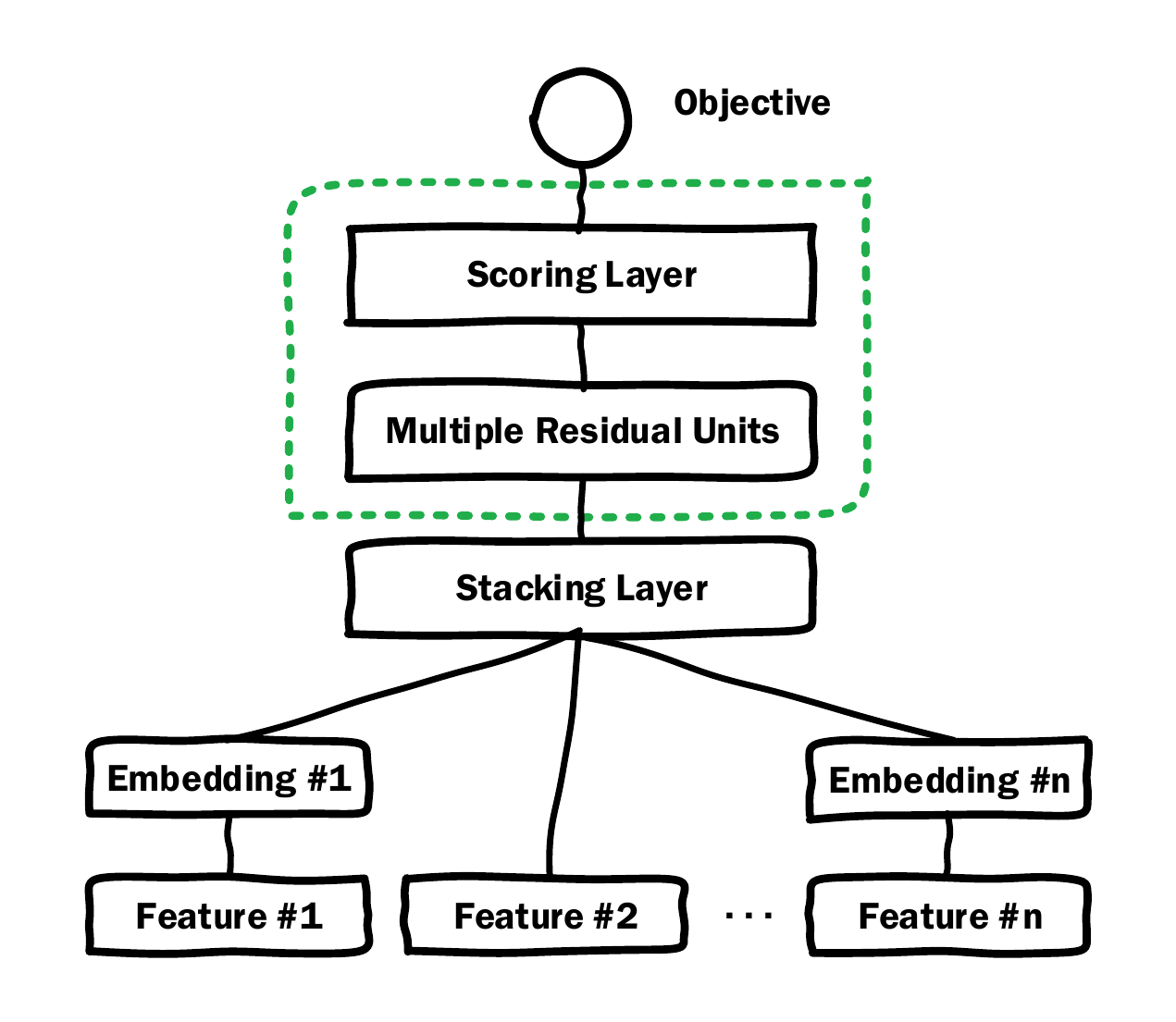}
	\caption{The Deep Crossing Model used to initialize the embedding layers of DEF}
	\label{fig:defdeepcrossing}
\end{figure}

The first step is to initialize $\{\mathbf{W}_j\}$, $\{\mathbf{b}_j\}$ in Equ.~\ref{equ:embedding} with the Deep Crossing model~\cite{ying2016DeepCrossing}\footnote{Code is available under https://github.com/Microsoft/CNTK/wiki/Deep-Crossing-on-CNTK}.  As can be seen from Fig.~\ref{fig:defdeepcrossing}, the embedding layers and the stacking layer are exactly the same as in the DEF model in Fig.~\ref{fig:defmodel}.  The difference is that the forest layer is replaced by the layers inside the dotted rectangle.  The \emph{multiple residual units} are constructed from the \emph{residual units}, which is the basic building block of the Residual Net~\cite{he2015deep} that claimed the world record in the ImageNet contest.  The use of residual units provides Deep Crossing with superior convergence property and better control to avoid overfitting.

While other DNNs can be applied in the context of DEF, Deep Crossing offers some unique benefits that especially suitable for this task. First of all, it is designed to enable large scale applications.  In ~\cite{ying2016DeepCrossing}, Deep Crossing was implemented on a multi-GPU platform powered by the open source tool called Computational Network Toolkit (CNTK)\cite{CNTK2014}\footnote{http://www.cntk.ai}.  The resulting system was able to train a Deep Crossing model with five layers of residual units ($10$ individual layers in total), with $2.2$ billion samples of more than $200 K$ dimensions.  The model significantly exceeded the offline AUC of a click prediction model in a major sponsored search engine.

Another key benefit of Deep Crossing is the ability to handle low level sparse features with high dimensions.  In the above example, three tri-letter grams with $50 K$ dimensions each were in the features list. The feature interactions (or cross features) were automatically captured among sparse features, and also w.r.t. other individual features such as CampaignID and  CampaignIDCount in Table~\ref{tab:ExampleOfIndividualFeatures}.  The resulting embedding features are dense, and in much lower dimensions that fall into the comfort zone of forest/tree-based models.

With the presence of a general \emph{scoring layer}, Deep Crossing also works with all the objective functions of DEF. 

It should be pointed out that there is no difference in the training process of using Deep Crossing as a stand-alone model versus DEF initialization.  In other words, the optimization is w.r.t. the embedding parameters, the parameters in the multiple residual units, and those in the scoring layer.  The difference is in the usage of the model after the training completes.  Unlike the stand-along model, DEF uses only the embedding parameters for subsequent steps.

\subsection{Initialize the Forest Layer with XGBoost and LightGBM}
\label{DEFStep2}
The embedding layers establish a forward function that maps the raw features into the stacking vector:

\begin{equation}
	\tilde{\mathbf{y}}_i = \mathcal{G}(\{\mathbf{x}_j\}_i;\{\mathbf{W}_j^0\}, \{\mathbf{b}_j^0\}),
\end{equation}
where $i$ is the index to the training samples.  The initial values of the embedding parameters are denoted as $\{\mathbf{W}_j^0\}$ and $\{\mathbf{b}_j^0\}$.  The nonlinear operator $\mathcal{G}$ is the combination of $g$ in Equ.~\ref{equ:embedding} and the vector-stacking operator in Equ.~\ref{equ:vectorstacking}. 

The second step in constructing a DEF builds on top of this function to initialize the forest layer.  This is accomplished by training a forest/tree-based model using the mapped sample and target pairs $\{(t_i, \tilde{\mathbf{y}_i})\}$, where $t_i$ is the target of the $i^{th}$ training sample.  Both XGBoost and LightGBM are applied to serve this purpose.

XGBoost~\cite{chen2016xgboost}\footnote{https://github.com/dmlc/xgboost} is a gradient boosting machine (GBM) that is widely used by data scientists, and is behind many winning solutions of various machine learning challenges.

LightGBM~\cite{meng2016communication}\footnote{https://github.com/Microsoft/LightGBM} is another open source GBM tool recently developed by Microsoft.  It uses histogram based algorithms to accelerate training process and reduce memory consumption~\cite{ranka1998clouds, jin2003communication} and also incorporates advanced network communication algorithms to optimize parallel learning.

The outcome of this step, either produced by XGBoost or LightGBM, becomes the initial values of the forest parameters including $\Psi^0$, $\Theta^0$, and $\Pi^0$.

\subsection{Joint Optimization with Partial Fuzzification}
\label{DEFStep3}
The third step is a joint optimization that refines the parameters of the embedding layers $\{\mathbf{W}_j\}$ and $\{\mathbf{b}_j\}$, and the parameters of the forest layer $\Theta$, and $\Pi$.  Note that the number and structure of trees (i.e., $\Psi$) are kept unchanged.  Ideally a joint optimization that solves these parameters holistically is preferred.  Unfortunately, this is non-trivial mostly due to the existence of the forest layer.  Specifically, the search of the best structure and the corresponding decision and weight parameters in the forest layer usually relies on greedy approaches, which are not compatible with the gradient-based search for the neural-based embedding layers.  To enable this, DEF has to overcome the hurdle of converting the refinement of the forest into a continuous optimization problem.  

We start by looking into the pioneer work in~\cite{suarez1999globally, richmond2015mapping, kontschieder2015deep}, where the above problem is solved by \emph{fuzzifying} the decision functions\footnote{Also referred as \emph{split function} or \emph{routing function} in literature} of the internal nodes.  Instead of making a binary decision on the $r^{th}$ internal node, a \emph{fuzzy split function} makes fuzzy decisions on each node.  The probability of directing a sample to its left or right child node is determined by a sigmoidal function:
\begin{align}
	\mu^{L}_r(\tilde{\mathbf{y}}) = \frac{1}{1+e^{-c_r(\mathbf{v}_r\cdot \tilde{\mathbf{y}}-a_r)}}, \nonumber\\
  	\mu^{R}_r(\tilde{\mathbf{y}}) = \frac{1}{1+e^{c_r(\mathbf{v}_r\cdot \tilde{\mathbf{y}}-a_r)}} \nonumber\\
    = 1-\mu^L_r(\tilde{\mathbf{y}}), \label{equ:fullfeaturesplit}
\end{align}
where $r$ denotes the index of the internal nodes, $\mathbf{v}_r$ is the weight vector, $a_r$ is the split value of the scalar variable $\mathbf{v_r}\cdot\tilde{\mathbf{y}}$, and $c_r$ is the inverse width that determines the fuzzy decision range. Outside this range, the assignment of a sample to the child nodes is approximately reduced to a binary split.  $\tilde{\mathbf{y}} \equiv \tilde{\mathbf{y}}_i$ is the stacking vector of the $i^{th}$ training sample.  The index $i$ is dropped here for simplicity.  The functions $\mu_r^L$, and $\mu_r^R$ are defined for the left child and the right child (if exists), respectively.  The prediction of the target $t$ is:
\begin{equation}
	\bar{t} = \sum_{l\in \mathcal{L}} \mu_l(\tilde{\mathbf{y}}) \pi_l,
	\label{equ:fulltreepredict}
\end{equation}
where $l$ is a leaf node in the set of all leaf nodes $\mathcal{L}$, $\mu_l(\cdot)$ is the probability of $\tilde{\mathbf{y}}$ landing in $l$, and $\pi_l$ is the corresponding prediction.  Note that $\mu_l(\cdot)$ is a function of all $\{\mu^{L,R}_r(\tilde{\mathbf{y}})\}$ in Equ.~\ref{equ:fullfeaturesplit} along the path from the root to $l$.

A direct benefit of such fuzzification is that a continuous loss function $Loss(\{t_i, \bar{t}_i\})$ is differentiable w.r.t. $\{\mathbf{v}_r\}$, $\{c_r\}$, $\{a_r\}$, and $\{\pi_l\}$.  The downside, however, is the cost of the prediction, which requires the traversal of the entire tree, as indicated in Equ.~\ref{equ:fulltreepredict}.  From Equ.~\ref{equ:fullfeaturesplit}, it can be seen that the split function depends on all dimensions of the stacking features.  As has been discussed in Sec.~\ref{subsec:defproperties}, this is also computationally expensive. 

The above approach will be referred as \emph{full fuzzification} hereafter, due to the fact that it requires full traverse of the forest, and has dependency on all feature dimensions.

DEF simplifies full fuzzification by having each internal node address only one dimension of the stacking vector that was selected by the forest/tree model to conduct the joint optimization.  The fuzzification on each node is simplified to the following:
\begin{align}
	\mu^{L}_r(\tilde{\mathbf{y}}) = \frac{1}{1+e^{-c_r(\tilde{y}^r-a_r)}}, \nonumber\\
  	\mu^{R}_r(\tilde{\mathbf{y}}) = \frac{1}{1+e^{c_r(\tilde{y}^r-a_r)}} \nonumber\\
    = 1-\mu^L_r(\tilde{\mathbf{y}}) \label{equ:partialfeaturepredictnode},
\end{align}
where $c_r$ is the inverse width, and $a_r$ is the split value on the $r^{th}$ node.  These parameters are initialized by the split function that has been learned in Sec.~\ref{DEFStep2}.  The binary decision is replaced by a fuzzy decision especially for the samples that land within the fuzzy decision region.  The joint optimization allows parameters $\{\mathbf{W}_j\}$, $\{\mathbf{b}_j\}$, $\Theta$, and $\Pi$ to evolve simultaneously to reach the model optimum. 

As compared with the split function in Equ.~\ref{equ:fullfeaturesplit}, the linear transform of the stacking vector $\tilde{\mathbf{y}}$ is removed.  This is because each node is dedicated to only one dimension of the stacking vector, which is denoted as $\tilde{y}^r$ in Equ.~\ref{equ:partialfeaturepredictnode}.  More specifically,  $\tilde{y}^r$ is the feature value of the dimension selected by the $r^{th}$ node based on $\Psi^0$ and $\Theta^0$. The prediction is the weighted average of all the leaf values of the tree, as shown in Equ.~\ref{equ:fulltreepredict}.  Compared to \emph{full fuzzification}, the time complexity is reduced by the length of the stacking vector, since the split on each node only relays on the dimension that is selected by the forest/tree model.

\subsection{Two-Step DEF vs. Three-Step DEF}
\label{sec:twostepandthreestep}
There are two options applying DEF.  The first option, referred hereafter as the Two-Step DEF, involves only the initialization steps in Sec.~\ref{DEFStep1} and Sec.~\ref{DEFStep2}.  Another option involves the full three steps including partial fuzzification described in the above section.  This option will be referred hereafter as the Three-Step DEF.

As mentioned in Sec.~\ref{subsec:defproperties}, the prediction time of the Two-Step DEF is proportional to $n_t d_t$, where $n_t$ is the number of trees, and $d_t$ is the average depth of the trees in the forest. For the Three-Step DEF, the partial fuzzification relies on information from all nodes.  As a result, the prediction time $T_2$ is a function of $n_t l_t$, where $l_t$ is the average number of nodes of the trees. The ratio of time complexity between Three-Step DEF and Two-Step DEF is $\frac{l_t}{d_t}$, which can grow rapidly as $d_t$ increases since $l_t$ can be roughly exponential in terms of $d_t$.
\section{Experiment Results}
\label{sec:experimentresults}
This section reports results using click prediction data.  The problem is to predict the probability of click on an ad (see context in Sec.~\ref{sponsored_search}), given input strings from a query, a keyword, an ad title, and a vector of dense features with several hundred dimensions.

As explained in Sec.~\ref{representation}, query strings are converted into tri-letter grams, and so are the keyword and title strings.  The dense features are mostly counting features similar to the \verb|CampaignIDCount| feature in Table~\ref{tab:ExampleOfIndividualFeatures}.  The raw input feature has around $150$K dimensions in total, and is a mix of both sparse and dense features.

All models including Deep Crossing model, XGBoost, and LightGBM share the same log loss function as the objective. The embedding vector is $128$ dimensional for each tri-letter gram and the dense features.  This leads to a stacking vector of $512$ dimensions.  Deep Crossing uses two residual units\footnote{See Sec.5.2 in the Deep Crossing paper~\cite{ying2016DeepCrossing} for more information about residual unit}, where the first is connected with the input stacking vector, and the second takes its output as the input.  The first residual unit has three layers with the sizes of $512$, $128$, and $512$, respectively.  The second residual unit has similar dimensions, except that the middle layer has only $64$ neurons.  

We used a simplified version of the Deep Crossing model that outperformed the production model in offline experiments, as reported in~\cite{ying2016DeepCrossing}. Its log loss performance is slightly worse than the original model but the model size is more suitable to demonstrate the efficacy of the DEF model.  As can be seen in the following experiments, DEF can reduce prediction time significantly even the Deep Crossing model is a simplified one.

Both XGBoost and LightGBM have parallel implementations.  We will use both of them to demonstrate that DEF works with different kinds of forest/tree-based models.  However, since parallel LightGBM is already set up in our environment, we will use it for most of the experiments.

In order to achieve a clean apple-to-apple comparison, the majority of the experiments are based on the prediction time using a single CPU processor, without relying on any of the SIMD instructions (referred hereafter as \emph{plain implementation}).

\subsection{Experiment with $3$ Million Samples: Log Loss Comparison}

\begin{table*}
\centering
		\begin{tabular}{ l | c | c | c | c }
			\hline			
			 & Deep Crossing & 2-Step (XGBoost) & 2-Step (LightGBM) & 3-Step (LightGBM)\\ \hline\hline
			Relative log loss & 100 & 99.96 & 99.94 &  99.81\\ \hline 
			Time (ms) & 2.272 & 0.168 & 0.204 & - \\ \hline 
			\hline  
		\end{tabular}	
	\caption{Comparison of performance and prediction time per sample between Deep Crossing and DEF.  The models are trained with $3.59$M samples.  Performance is measured by a relative log loss using Deep Crossing as the baseline.  The prediction time per sample was measured using $1$ CPU processor}
	\label{tab:table-3m}
\end{table*}
For the first set of experiments, the training data has $3.59$M (million) samples, and the test data has $3.64$M samples. The Deep Crossing model was trained on $1$ GPU and converged after $22$ epoches.  A stacking vector was generated for each sample based on the forward computation of the Deep Crossing model, and was used as the input to train the forest models including both XGBoost and LightGBM.  The resulting models are the Two-Step DEFs.  We also experimented with the partial fuzzification model initialized by LightGBM. In all the experiments hereafter, the performance will be reported in terms of relative log loss, defined as the following:
\begin{equation}
	\gamma_r^{DEF} = \frac{\gamma^{DEF}}{\gamma^{DC}} \times 100,
	\label{equ:relativelogloss}
\end{equation}
where $\gamma$ and $\gamma_r$ are the actual and relative log losses, respectively, and $DC$ represents the Deep Crossing Model for simplicity.  As with the actual log loss, a smaller relative log loss indicates better performance.

\subsubsection{Two-Step DEF with XGBoost as the Forest Layer}
XGBoost converged after $1108$ iterations with a maximum depth of $7$.  The Two-Step DEF slightly outperformed Deep Crossing, as shown in Table~\ref{tab:table-3m}. 

\subsubsection{Two-Step DEF with LightGBM as the Forest Layer}
LightGBM model converged after $678$ iterations with a maximum number of leaves of $128$.  As shown in Table~\ref{tab:table-3m}, the Two-Step DEF using LightGBM performed better than Deep Crossing in log loss.  Note that the apple-to-apple performance comparison between XGBoost and LightGBM requires a different setting, which is less of interest in this experiment.
\subsubsection{Three-Step DEF with Partial Fuzzification}
Taking the LightGBM model as the initial forest layer, the joint optimization using partial fuzzification achieved slightly better accuracy with a relative log loss of $99.81$, as shown in Table~\ref{tab:table-3m}.

\begin{figure}
	\centering
		\includegraphics[scale=0.5]{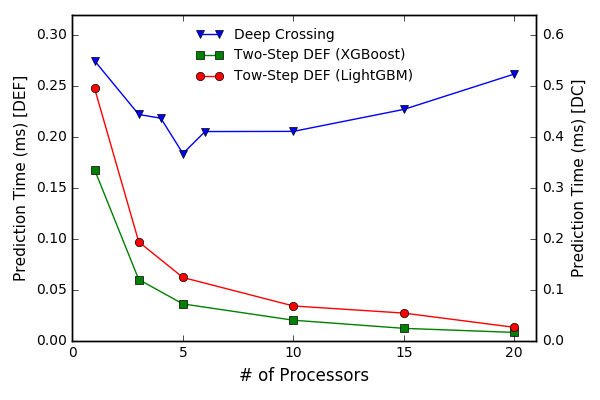}
	\caption{The Comparison of scalability of the prediction runtime per sample on CPU processors among Deep Crossing, Two-Step DEF with XGBoost and Two-Step DEF with LightGBM.  Note that the axis of prediction time for DEF is on the left side, while the one for Deep Crossing is on the right side}
	\label{fig:timevscore}
\end{figure}

\subsection{Experiment with $3$ Million Samples: Prediction Time Comparison}
As shown in Table ~\ref{tab:table-3m}, the prediction time for Deep Crossing, Two-Step DEF with XGBoost, and Two-Step DEF with LightGBM are $2.272$ ms, $0.168$ ms, and $0.204$ ms, respectively.  The prediction time is measured at the per sample level using one CPU processor, with all the I/O processes excluded.

We also experimented the prediction time with different number of processors.  As shown in Fig.~\ref{fig:timevscore}, the prediction time for DEF with both XGBoost and LightGBM decreases as the number of processors increases.  The prediction time for Deep Crossing, on the other hand, started to increase after reaching a minimum of $0.368$ ms with five processors.  This is expected because of its inherently sequential computation between the consecutive layers. With more than five processors, the parallelism is outweighed by the communication overhead.

A caveat to call out is that both XGBoost and LightGBM are using sample partition for parallelization in the current implementation of the predictor. This is in theory different from the per sample prediction time that we want to measure.  However, since the forest/tree-based models are \emph{perfectly parallel}, we expect that the effect of sample partition will be a good approximation of partitioning with individual trees.

The Deep Crossing predictor in this experiment was implemented with the Intel MKL to utilize multiple processors.  This implementation is faster than the plain implementation even when there is only one processor involved.  As a result, the prediction time in this section can not be compared with those in other experiments.

Note that the speedup was achieved with DEF performing better in terms of log loss.  If the goal is to achieve on-par performance in practice, the number of iterations (trees) can be significantly reduced to further reduce the run-time latency.

Also note that we didn't compare the prediction time of the Three-Step DEF in Table~\ref{tab:table-3m}. This is because the Three-Step DEF runs on GPUs, as detailed in the Appendix.  From the discussion in Sec.~\ref{sec:twostepandthreestep}, it is clear that the Three-Step DEF is always slower than the Two-Step DEF, if both measured against a CPU processor.

\subsection{Experiment with $60$ Million Samples}
The same experiment was applied to a dataset which contains 60M samples. The test data has $3.59$M samples.  The comparison of the performance and speed is shown in Table~\ref{tab:table-60m}.  The Deep Crossing model converged after $14$ epochs, of which the log loss is normalized to $100$. The stacking vector generated by the Deep Crossing model was then used to train LightGBM, which converged with a normalized log loss of $99.77$ after $170$ iterations. The LightGBM in this experiment used a maximum number of leaves of $2048$.  The prediction time per sample is $2.272$ ms for the Deep Crossing Model, and is $0.162$ ms for DEF.  The joint optimization was not conducted because the training process took too long.

\begin{table}
\centering
		\begin{tabular}{ l | c | c }
			\hline			
			 & Deep Crossing & 2-Step DEF\\ \hline\hline
			Relative log loss & 100 & 99.77 \\ \hline 
			Time (ms) & 2.272 & 0.162 \\ \hline 
			\hline  
		\end{tabular}	
	\caption{Comparison of performance and prediction time per sample between the Deep Crossing and Two-Step DEF using LightGBM that are trained with $60$M samples.  The predictions both were conducted on $1$ CPU processor.  Performance is measured by relative log loss using Deep Crossing as the baseline}
	\label{tab:table-60m}
\end{table}



\subsection{Experiment with $1$ Billion Samples}
This experiment applied DEF to a large sample set with $1$B (billion) samples.  A distributed LightGBM was used to train the forest layer on a $8$ server cluster.  We kept the test data the same as before.  The Deep Crossing model was trained on a GPU cluster with $16$ GPUs, and converged after $30$ epochs.    At the time this paper is submitted, the best LightGBM model has 70 trees with the maximum number of leaves equal to $16384$.  As shown in Table~\ref{tab:table-1b},  the LightGBM model performed slightly worse than the Deep Crossing model.  Regarding the prediction speed, LightGBM outperformed Deep Crossing approximately by more than $10$ times.

\begin{table}
\centering
		\begin{tabular}{ l | c | c }
			\hline			
			 & Deep Crossing & 2-Step DEF\\ \hline\hline
			Relative log loss & 100 & 100.55 \\ \hline 
			Time (ms) & 2.272 & 0.204 \\ \hline 
			\hline  
		\end{tabular}	
	\caption{The same as Table~\ref{tab:table-60m} but for the models trained by $1$B samples}
	\label{tab:table-1b}
\end{table}


\begin{table}
\centering
		\begin{tabular}{ l | c | c | c}
			\hline			
			 & $3$M & $60$M & $1$B \\ \hline\hline
			Deep Crossing & 100 & 99.78 & 98.71\\ \hline 
			2-Step DEF (LightGBM) & 99.94 & 99.56 & 99.24\\ \hline 
			\hline  
		\end{tabular}	
	\caption{Comparison of performance of Deep Crossing and Two-Step DEF with LightGBM in the $3$M, $60$M and $1$B experiments.  Performance is measured by a relative log loss using Deep Crossing in the $3$M experiment as the baseline}
	\label{tab:logloss3experiments}
\end{table}

\subsection{The Effect of Scale}
Table~\ref{tab:logloss3experiments} compares log loss performance of the Deep Crossing models and the DEF models as the number of samples increases.  The baseline log loss is based on the Deep Crossing model trained with $3.59$M samples.  As can be seen from the table, the log loss decreases as the number of samples increases to $60$M and $1$B.  DEF models exhibit the same trend. It is important to point out that even though the DEF model trained at $1$B samples performed worse than the Deep Crossing counterpart, it is still better than the Deep Crossing model trained with $60$M samples.

\section{Related Work}
\label{sec:relatedwork}
Combining a decision forest/tree with DNN has been studied before.  Deep Neural Decision Forest~\cite{kontschieder2015deep} represents the latest result along this direction, where a decision forest is generalized and trained together with DNN layers with Back Propagation.  It avoids the hard problem of explicitly finding the forest structure $\Psi$ but has to rely on decision nodes that are the functions of all input dimensions.  As a result, the run-time complexity is $O(n_t l_t n_f)$ instead of $O(n_t d_t)$\footnote{See Sec.~\ref{subsec:defproperties} for definitions}.  The work in~\cite{richmond2015mapping} learns a forest first and maps it to a DNN to provide better initialization when training data is limited.  The DNN is then mapped back to a forest to speed up run-time performance.  In order to achieve $O(n_t d_t)$ complexity, it has to use an approximation to constrain the number of input nodes (the features) to the decision nodes, which significantly deteriorates the accuracy.  In contrast, DEF without joint optimization achieves $O(n_t d_t)$ run-time complexity with on-par or slightly better accuracy than the DNN counterpart.

The idea of pre-training components of the network separately as initialization (similar to Sec. ~\ref{DEFStep1}) in order to construct high level features has also been investigated in autoencoders and Deep Belief Networks (DBNs) ~\cite{bengio2009learning, vincent2010stacked}. Nevertheless, this paper is fundamentally different from those previous works in two main aspects. First, the pre-training in autoencoders and DBNs are performed in unsupervised fashion as opposed to Sec. ~\ref{DEFStep1} in which supervised Deep Crossing technique is used. Secondly, our work is mainly motivated by obtaining efficient prediction while maintaining high performance, whereas the principal motivation for pre-training in those previous papers is mostly to achieve high-level abstraction.

As discussed earlier, the inefficient serving time of DNNs is mainly because of 
expensive run-time computation of multiple layers of dense matrices. Therefore, a natural idea to obtain efficiency in online prediction might be to convert DNNs into an (almost) equivalent shallow neural network once they are trained. This idea even sounds theoretically feasible due to the richness of neural networks with simply one single hidden layer according to Universal Approximation Theorem~\cite{hornik1989multilayer}. Nevertheless, the function modeled 
by DNN is fairly complicated due to deep embedding features. Thus, despite universality, a single layer neural network will become impractical as it may need astronomical number of neurons to model the desired function~\cite{cybenko1989approximation}.

It is also worthwhile to mention the work in~\cite{he2014practical}, which combines a boosted decision forest with a sparse linear classifier.  It was observed that the combined model performs better than either of the models on its own.  If this is the case, DEF can be easily extended to add a linear layer on top of the forest layer for better performance.  The computational overhead will be negligible.

DEF is built on top of the Deep Crossing model~\cite{ying2016DeepCrossing}, the XGBoost~\cite{chen2016xgboost} model, and the LightGBM~\cite{meng2016communication} model.  While the deep layers in the Deep Crossing model are replaced with the forest/tree layer in the DEF model, they are critical in discovering the discriminative embedding features that make the construction of the forest/tree models an easier task.

\section{Conclusion and Future Work}
\label{sec:conclusionfuturework}
DEF is an end-to-end machine learning solution from training to serving for large scale applications using conventional hardware.  Built on top of the latest advance in both DNN and forest/tree-based models, DEF handles low-level and hight dimensional heterogeneous features like DNN, and serves the high precision model like a decision tree.

DEF demonstrates that a simple two-step approach can produce near optimal results as more complex joint optimization through fuzzification.  While DNNs with fuzzification have been reported to perform better than the plain DNNs, DEF is the first we are aware of that outperforms the DNN counterpart without compromising the property of fast serving.

Also because of the two-step approach, DEF serves as a flexible framework that can be adapted to different application scenarios.  In the paper, we demonstrated that DEF with XGBoost works as well as DEF with LightGBM.  As a result, applications can choose the forest/tree-based models that work best in terms of availability and performance.  For instance, we decided to use LightGBM (instead of XGBoost) for large scale experiments because we had it set up running in a parallel environment.

DEF can work with other types of DNN-based embedding.  However, embedding approaches such as LSTM and CNN cost significant amount of run-time calculation that cannot be saved by the forest/tree layer.  This is a direction we are looking into as a future work.
\section{Acknowledgments}
\label{sec:acknowledgements}
The authors would like to thank Jian Jiao (Bing Ads), Qiang Lou (Bing Ads), T.~Ryan Hoens, Ningyi Xu (MSRA), Taifeng Wang (MSRA), and Guolin Ke (MSRA) for their support and discussions that benefited the development of DEF.

\begin{appendices}
\section{Implementation of partial fuzzification in DEF}
\label{sec:implementationfuzzification}
The joint optimization using partial fuzzification in DEF was developed as a forest node in CNTK so that it is compatible with the embedding and scoring layers of Deep Crossing.  Both the forward and backward operations are implemented on GPU for acceleration.
 
\subsection{Initialization}
The structure of the trees and the forest parameters $\{c_r\}$, $\{a_r\}$ and $\{\pi_l\}$ are initialized with the trained forest/tree model.  The embedding parameters $\{\mathbf{W}_j\}$, $\{\mathbf{b}_j\}$ are initialized by the trained Deep Crossing model.

\subsection{Forward Propagation}
The prediction of a sample on leaf $l$ is defined as 
\begin{equation}
	t_l = \pi_l\prod_{r\in \Omega^l} \mu_r^{L, R}(\tilde{\mathbf{y}}),
	\label{equ:fulltreepredictleaf}
\end{equation}
where $r$ and $l$ denote the index of internal and leaf nodes respectively, $\Omega^l$ represents the set of internal nodes along the route from root to leaf $l$.  The stacking vector $\tilde{\mathbf{y}}$ is computed by Equ.~\ref{equ:embedding}.  The terms $\mu_l^{L,R}(\tilde{\mathbf{y}})$ are defined in Equ.~\ref{equ:partialfeaturepredictnode}.  The selection of $\mu_l^L(\tilde{\mathbf{y}})$ or $\mu_l^R(\tilde{\mathbf{y}})$ depends on which child node $r$ directs to along the path.  The raw score $\bar{t}$ is defined as:
\begin{equation}
	\bar{t} = \sum_{l\in \mathcal{L}} t_l,
	\label{equ:fulltreepredictsum}
\end{equation}
where $\mathcal{L}$ represents the set of all the leaf nodes in the forest.  The raw score is then transformed to the prediction score via a sigmoidal function.  The prediction score is a function of the split parameters $\{c_r\}$ and $\{a_r\}$, the leaf parameter $\{\pi_l\}$, and the stacking vector $\tilde{\mathbf{y}}$. 

\subsection{Backward Propagation}
The backward propagation refines the fuzzification parameters $\{c_r\}$, $\{a_r\}$ and $\{\pi_l\}$, and the input stacking vector $\tilde{\mathbf{y}}$.  Define the gradient of the objective function (denoted as $\mathcal{O}$) with respect to the raw score $\bar{t}$ as:
\begin{equation}
	\delta_t = \frac{\partial \mathcal{O}}{\partial \bar{t}}.
	\label{equ:gradientrawscore}
\end{equation}
The updating equation for optimizing $\mathcal{O}$ w.r.t. the prediction parameter on leaf $l$ (i.e., $\pi_l$) is:
\begin{equation}
	\frac{\partial \mathcal{O}}{\partial \pi_l} = \frac{\partial \mathcal{O}}{\partial \bar{t}} \frac{\partial \bar{t}}{\partial t_l} \frac{\partial t_l}{\partial \pi_l} = \delta_t\prod_{r\in \Omega^l} \mu_r^{L, R}(\tilde{\mathbf{y}}).
	\label{equ:pigradient}
\end{equation}
The gradient w.r.t. $c_r$ on the $r$th internal node can be expressed as:
\begin{equation}
	\frac{\partial \mathcal{O}}{\partial c_r}= \frac{\partial \mathcal{O}}{\partial \bar{t}} \sum_{l \in \mathcal{L}^r} \frac{\partial \bar{t}}{\partial t_l} \frac{\partial t_l}{\partial c_r} = \sigma_t \sum_{l \in \mathcal{L}^r} \pi_l \cdot \eta_l \prod_{p\in \Omega^l, p \ne r} \mu_p^{L, R}(\tilde{\mathbf{y}}),
	\label{equ:cgradient}
\end{equation}
where $p$ is the index of the internal nodes, and $\mathcal{L}^r$ is the set of the leaf nodes from which the paths to root contain the internal node $r$.  Depending on which child the $r$th node heads to along the path from root to leaf $l$, $\eta_l$ is defined as:
\begin{align}
	\eta_l^L = \mu_r^{L}(\tilde{\mathbf{y}}) (1-\mu_r^{L}(\tilde{\mathbf{y}}) (\tilde{y}^r-a_r),\nonumber \\
    \eta_l^R = \mu_r^{R}(\tilde{\mathbf{y}}) (1-\mu_r^{R}(\tilde{\mathbf{y}}) (a_r - \tilde{y}^r).
	\label{equ:eta}
\end{align}
The gradient of $\mathcal{O}$ w.r.t. $a_r$ has a similar expression as $c_r$ in Equ.~\ref{equ:cgradient}: 
\begin{equation}
	\frac{\partial \mathcal{O}}{\partial a_r}= \frac{\partial \mathcal{O}}{\partial \bar{t}} \sum_{l \in \mathcal{L}^r} \frac{\partial \bar{t}}{\partial t_l} \frac{\partial t_l}{\partial a_r} = \sigma_t \sum_{l \in \mathcal{L}^r} \pi_l \cdot \zeta_l \prod_{p\in \Omega^l, p \ne r} \mu_p^{L, R}(\tilde{\mathbf{y}}),
	\label{equ:agradient}
\end{equation}
where $\zeta_l$ is defined as:
\begin{align}
	\zeta_l^L = -c_r \cdot \mu_r^{L}(\tilde{\mathbf{y}}) (1-\mu_r^{L}(\tilde{\mathbf{y}}) , \nonumber \\
    \zeta_l^R = c_r \cdot \mu_r^{R}(\tilde{\mathbf{y}}) (1-\mu_r^{R}(\tilde{\mathbf{y}}).
	\label{equ:zeta}
\end{align}
Here, once again, the selection of expression of $\zeta_l$ is path-dependent.  For the input vector $\tilde{\mathbf{y}}$ on dimension $k$ (denoted as $\tilde{y_k}$), the gradient is written as:
\begin{align}
	\begin{split}
	\frac{\partial \mathcal{O}}{\partial \tilde{y}_k} &= \frac{\partial \mathcal{O}}{\partial \bar{t}} \sum_{l \in \mathcal{L}^k} \frac{\partial \bar{t}}{\partial t_l} \frac{\partial t_l}{\partial \tilde{y}_k} \\ 
    &= \sigma_t \sum_{l \in \mathcal{L}^k} \pi_l \sum_{q \in \Omega^{l,k}} \xi_{l,q} \prod_{p\in \Omega^l, p \ne q} \mu_p^{L, R}(\tilde{\mathbf{y}}),
	\end{split}
    \label{equ:ygradient}
\end{align}
where $q$ is the index of the internal nodes, $\mathcal{L}^k$ is the set of leaf nodes to which the routes from root pass one or more internal nodes with a split feature on $\tilde{y}_k$.  The set $\Omega_{l,k}$ is the internal nodes that dedicate to the $k$th dimension of $\tilde{\mathbf{y}}$ along the route to leaf $l$, and the set $\Omega_l$ is the same as in Equ.~\ref{equ:fulltreepredictleaf}. The term $\xi_{l, q}$ is defined as:
\begin{align}
	\xi_{l, q}^L = c_q \cdot \mu_q^{L}(\tilde{\mathbf{y}}) (1-\mu_q^{L}(\tilde{\mathbf{y}}) , \nonumber \\
    \zeta_{l, q}^R = -c_q \cdot \mu_q^{R}(\tilde{\mathbf{y}}) (1-\mu_q^{R}(\tilde{\mathbf{y}}).
	\label{equ:xi}
\end{align}
As $\tilde{{\mathbf{y}}}$ is a function of $\{\mathbf{W}_j\}$, $\{\mathbf{b}_j\}$, the gradient of $\mathcal{O}$ with respect to the two embedding parameters can be computed through the chain rule using $\frac{\partial \mathcal{O}}{\partial \tilde{y}_k}$.    
\end{appendices}

\bibliographystyle{acm}
\bibliography{DeepEmbeddingForest}

\end{document}